\documentclass[11pt]{article}
\usepackage{geometry}
\usepackage{coling2020}
\usepackage{times}
\usepackage{url}
\usepackage{latexsym}
\usepackage{microtype}
\usepackage{graphicx}
\usepackage{url}
\usepackage{float}
\usepackage{makecell}
\usepackage[utf8]{inputenc}
\usepackage[arabic, english]{babel}
\usepackage{adjustbox}

\colingfinalcopy 

\title{KEIS@JUST at SemEval-2020 Task 12: Identifying Multilingual Offensive Tweets Using Weighted Ensemble and Fine-Tuned BERT}

\author{Saja Khaled Tawalbeh, Mahmoud Hammad, Mohammad AL-Smadi\\Computer Science Department\\ 
Jordan University of Science and Technology\\ 
P.O.Box: 3030 Irbid 22110, Jordan. \\
{\tt sajatawalbeh91@gmail.com, \{m-hammad, masmadi\}@just.edu.jo}}

\date{}

\begin{document}
\maketitle
\begin{abstract}
This research presents our team KEIS@JUST participation at SemEval-2020 Task 12 which represents shared task on multilingual offensive language. We participated in all the provided languages for all subtasks except sub-task-A for the English language. Two main approaches have been developed the first is performed to tackle both languages Arabic and English, a weighted ensemble consists of Bi-GRU and CNN followed by Gaussian noise and global pooling layer multiplied by weights to improve the overall performance. The second is performed for other languages, a transfer learning from BERT beside the recurrent neural networks such as Bi-LSTM and Bi-GRU followed by a global average pooling layer. Word embedding and contextual embedding have been used as features, moreover, data augmentation has been used only for the Arabic language.
\end{abstract}

\section{Introduction}
\label{intro}
Natural language processing field has the researchers' attention especially with the rapid use of social media sites, for instance, Twitter, Facebook, YouTube comments, and macro blogs. Consequently, offensive, aggressive and hate-speech language identification problems that perform the  automatic detection of these  problems from textual data. Moreover, the main motivation to reduce the behavior of hate speech and offensive/aggressive language on user attitude and content, in particular, on social media.\\

The offensive detection in Arabic social media is a serious task. This refers to the Arabic language contains violent words represent both violent context and not a violent context besides Arabic is dialects language \cite{elfardy2013sentence}. For instance, the word Killing in Arabic meaning represents a violent meaning and not a violent meaning in different contexts appears in \cite{alhelbawy2016towards} tweets. The overall studies that detected on the offensive language have been applied to the English language. However, the research that regards in the Arabic language in this domain of NLP applications has been restricted according to the lack of the resources that tackle the same issue compared to the English language \cite{mubarak2017abusive,abozinadah2015detection}. The researches that had been conducted in Arabic evaluated on small datasets collected from Twitter API such as \newcite{mubarak2017abusive} has been evaluated their proposed approach on 1100 annotated tweets. \\

In this research, we describe our participation team KEIS@JUST at SemEval-2020 Task 12 which describes the shared task on offensive language as a multilingual shared task ({\em i.e.} Arabic, English, Danish, Greek, and Turkish). Moreover, We participated in all languages for the provided subtasks except sub-task-A for the English language. Two approaches have been implemented aim to solve the shared task, the first is performed to tackle both languages Arabic and English, a weighted ensemble consists of Bi-GRU and CNN followed by Gaussian noise and global pooling layer multiplied by weights to improve the overall performance. Consequently, the implemented approach performed to solve sub-task-A (offensive language identification), sub-task-B (automatic categorization), and sub-task-C (offense target identification). The second performed for other languages, a state of art transfer learning from BERT embedding multi\_cased 12A pre-trained model besides the recurrent neural networks such as Bi-LSTM and Bi-GRU followed by a global average pooling layer. Consequently, the implemented approach performed to solve sub-task-A (offensive language identification). Word embedding and contextual embedding have been used as features, moreover, data augmentation has been used only for the Arabic language and we rely on the AraVec embedding \cite{aravec} for data augmentation which aims to create more dataset that helps to train the model. To evaluate our results OffensEval 2020 \cite{2020-semeval} provided multilingual Dataset. The best results for the KEIS@JUST team ranked 11th place out of 56 teams with 86.55\% F1-macro in the Arabic language, ranked 12th place out of 39 teams with 76.1\% F1-macro in the Danish language, ranked 28th place out of 37 teams with 76.1\% F1-macro in the Greek language, ranked 32th place out of 46 teams with 73.3\% F1-macro in the Greek language.

\section{Related Work}
\label{related:pdf}
Offensive content on social media has recent attention \cite{schmidt2017survey,founta2018large,malmasi2018challenges} according to the negative effects on its users, for instance, demeaning comments or hate speech utterance. The offensive language detection on Arabic social media users considered an important step to prevent social society from these negative effects.\\

Several of previous researches have been presented comprehensive studies which tend to describe the main key of the proposed task \newcite{schmidt2017survey}, and \cite{fortuna2018survey}, moreover \cite{davidson2017automated} presents dataset for hate speech detection, \cite{trac2018report} presents dataset for aggressive language, and \cite{OLID} presents OLID dataset for the previous shared task of offensive language. Additionally, \cite{spertus1997smokey} shows the earliest efforts in hate speech detection that performs a decision tree-based classifier. Moreover, Offensive identification for sentences have been tried for several languages behind the English such that, Arabic \newcite{mubarak2017abusive} and \cite{al2019detection}, German \cite{ross2017measuring,fivser2017legal,su2017rephrasing}.\\

There are lack of researches in the offensive language for Arabic research community, for instance, \cite{abdelfatah2017unsupervised} introduced k-means for violence utterance on twitter. MADIMARA has been used to extract morphological features as well as they used TF-IDF to represent dataset on the vector space model. \cite{malmasi2017detecting} presented system to detect hate speech using lexical features and a linear SVM classifier depending on n-grams. Similarity, \cite{alakrot2018towards} introduced SVM classifier trained on word-level features. N-grams and stemming used as features. \cite{mulki2019hsab} proposed L-HSAB the first dataset for hate speech and abusive language. The dataset collected from twitter API focusing on Syrian and Lebanese tweets rich of toxic utterance. The dataset has been trained on Naive Bayes classifier.\\ 

For English language, several reseacrchers used transformers ,for instance, \cite{liu-etal-2019-nuli} Proposed a fine-tuned technique for the Bidirectional Encoder Representation from Transformer (BERT) with word unigrams, word2vec, and Hatebase have been used as features. Similarity, \cite{zhu-etal-2019-um} Introduced a fine-tuned a BERT based classifier depends on linear SVM trained on character n-gram as a feature. \cite{pelicon-etal-2019-embeddia} Proposed a fine-tuned a BERT and LSTM neural network architecture with automatically and manually crafted features were used namely: word embedding, TFIDF, POS sequences, BOW, the length of the longest punctuation sequence, and the sentiment of the tweets features. However, several researchers applied machine and deep learning, for instance,
\cite{mahata-etal-2019-midas} Proposed an ensemble technique consist of Convolutional Neural Network, Bidirectional LSTM with attention, and Bidirectional LSTM + Bidirectional GRU. \cite{han-etal-2019-jhan014} Presented two approaches namely: bidirectional with GRU and probabilistic model modified sentence offensiveness calculation (MSOC) trained using word2vec embedding.

\section{Methodology}
\label{System:pdf}
Shared task on Multilingual Offensive Language Identification in Social Media (OffensEval 2020) \cite{2020-semeval}, in this section, we will describe the shared task and the implemented system.

\subsection{Task Description}
OffensEval 2020 \newcite{2020-semeval} the shared task on multilingual offensive language \cite{sigurbergsson2020offensive}, \cite{mubarak2020arabic}, \newcite{pitenis2020offensive}, \cite{rosenthal2020largescale}, \cite{rosenthal2020}, and \cite{coltekikin2020}, however, the first offensive language task was organized at \cite{zampieri2019semeval}, \newcite{OLID}, and the recent aggressive multilingual language proposed by \cite{trac2020misogyny}. The shared task consist of three sub-tasks the goal of each sub-task represents as the following:
\subsubsection{Sub-task A}
Offensive language identification, aims to identify whether a tweet contains a non-acceptable language (profanity) or an offensive content. Moreover, sub-task A is a multilingual sub-task for five languages namely: Arabic, English, Danish, Greek,and Turkish. This sub-task is a binary classification, where each tweet has a labeled offensive (OFF) or not offensive (NOT). Our team (KEIS@JUST) participate in all languages for sub-task A except sub-task A for English language. 

\subsubsection{Sub-task B}
Automatic categorization of offense types, aims to identify whether an offense tweet contains targeted or non-targeted profanity and swearing. This sub-task is a binary classification provided only for English language, where each tweet has a labeled targeted (TIN) or untargeted (UNT).

\subsubsection{Sub-task C}
Offense target identification, aims to determine whether the offense target of the tweet is one of three tags namely: an individual (IND), group (GRP) or other (OTH). Other contains several tags ({\em i.g.} a situation, an organization, an event, or an issue). This sub-task provided only for English language.

\subsection{KEIS@JUST System}
\label{ssec1:layout}
\subsubsection{Weighted Ensemble (KEIS-BiGRUCNN)}
\label{Weighted:layout}
The main intuition of ensemble that combining the predictions comes from \cite{yu1977exploratory} that combines two regression techniques, after that \cite{dasarathy1979composite} had been presented the combination of two or more models. In this research, we will perform weighted ensemble technique consist of two models namely: KEIS-BiGRU and KEIS-CNN. The following provides more details about the implemented techniques.

\begin{figure*}
\centering
  \includegraphics[width=1\textwidth]{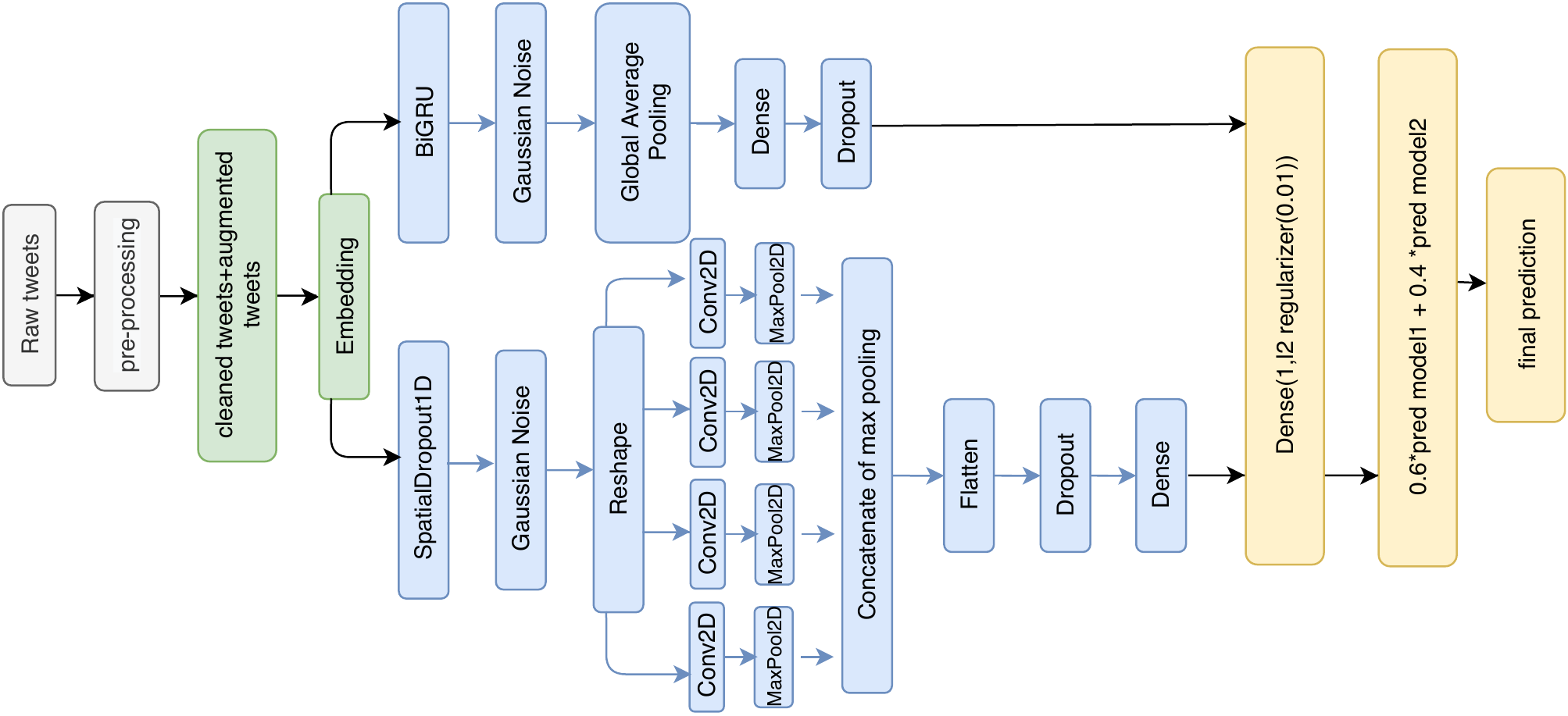}
  \caption{The implemented architecture of KEIS-BIGRU-CNN model}
  \label{fig:BiGRUcnn_fig}
\end{figure*}

\begin{itemize}
\item \textbf{Bidirectional-GRU (KEIS-BiGRU):} \\ Recurrent Neural Network (RNN) suffers from a gradient vanishing problem. Long Short Term Memory (LSTM) \newcite{lstm} has been proposed to solve the mentioned problem. Gated Recurrent Unit (GRU) \cite{chung2014empirical} as well as have been proposed to solve the gradient vanishing problem. Two gates have been used (reset and update gate) in GRU architecture. The main reasons that prefers  applied GRU over LSTMs (i.e. GRU tends to provide parallel performance with less complex structures compared to LSTM). The mathematical formulation of GRU can be expressed as:

\begin{equation}
 h_{t} = (1 - z_{t}) \odot h_{t-1} + z_{t} \odot \tilde h_{t}  \label{eq:sep_var} 
\end{equation}

where
\begin{equation}
\tilde h_{t} = tanh(W_{h}x_{t} + r_{t} \odot (U_{h}h_{t-1}) + b_{h}) \label{eq:HT}
\end{equation}

\begin{equation}
z_{t} = \sigma(W_{z}x_{t} + U_{z}h_{t-1} + b_{h}) \label{eq:ZT}
\end{equation}

\begin{equation}
r_{t} = \sigma(W_{r}x_{t} + U_{r}h_{t-1} + b_{h}) \label{eq:RT}
\end{equation}

where \(z_{t}\) EQ (\ref{eq:ZT}) explains the update gate, \(r_{t}\) EQ (\ref{eq:RT}) represent the reset gate, and \(\tilde h_{t}\) EQ (\ref{eq:HT}) explains the current memory content. Weight matrices are represented as \(W_h, W_z, W_r, U_h\)  \(U_z, U_r, b_h \), \(x_t\) explains the vector input to the timestep \(t\), \(ht\) EQ (\ref{eq:sep_var}) explains the final current exposed hidden state, and \(\odot\) explains the element-wise multiplication.\\

It's started with passing a sequence of words through an embedding layer followed Bidirectional GRU layer of 128 neurons, then Gaussian Noise of (0.1). Afterword, Global Average Pooling has been used to extract the discriminative features of the input tweet to prepare that for the next layer. Dense layer of 35 neurons has been applied followed by Dropout layer of (0.2) to prevent overfitting. Finally, the output layer will be Dense of (1) neuron with sigmoid function.

\item \textbf{Convolutional Neural Networks (KEIS-CNN):} \\
In deep learning, Convolutional Neural Networks (CNN) show the significant success in the fields of computer vision. The first CNN architecture for text classification proposed by \cite{kimconvolutional} which provides a remarkable enhancement on the performance of NLP tasks. Consequently, it can obtain the linguistic patterns from window of sequence words represented as embedding vectors. \\

It's started with passing a sequence of words through an embedding layer followed by Gaussian Noise of (0.1). As we know Conv2D have to reshape the input  to be compatible to receive the previous shape. Four Conv2D layers have been used. Each individual Conv2D sharing different filter size (1, 3, 5, 7) respectively. Moreover, the number of filters is 36 for all layers which aims to obtain the local information features. Afterward, each Conv2D layer passed to Max Pool 2D layer (MaxPool2D). In the last step, each layer has been concatenated together which aims to identify better output. In order to feed the next layer, we used Dropout layer of (0.25) to reduce over-fitting followed by dense layers of 35 neurons. The output is fed into single sigmoid which can obtain the output class of the given tweet.
\end{itemize}
For training step, the implemented KEIS-BiGRUCNN ensemble approach applied \newcite{aravec} embedding as pre trained model with 300 dimensions for Arabic language that prepared for training step. In contrast, we applied word2vec embedding for English proposed by \cite{mikolov2013distributed}, the pre trained embedding avalible at github acount \footnote{\url{https://github.com/felipebravom/AffectiveTweets/releases/download/1.0.0/w2v.twitter.edinburgh10M.400d.csv.gz}} with 400 dimensions. Several hyper-parameter have been used for optimization. Table \ref{parameter} provides more details about the value of each parameter have been used during the training step for both approaches. It's worth mentioning that amsgrad optimizer proposed by \cite{tran2019convergence} the updated version of adam optimizer with slight enhancement regards to the system performance.  The final step, after we obtained the final prediction for each model, the predictions have been multiplied by the best chosen weight to enhance the over all results. The ensemble architecture is shown in Fig. \ref{fig:BiGRUcnn_fig} KEIS-BiGRUCNN has been used to solve sub-task A for Arabic language and sub-task B,C for English language.
\begin{table}[t!]
\begin{center}
\begin{tabular}{|l|l|}
\hline \bf parameter & \bf Value \\ \hline
number of epochs & 20  \\
batch size & 128  \\
optimizer & amsgrad  \\
learning rate & 0.01  \\
weight for KEIS-CNN & 0.4  \\
weight for KEIS-BiGRU & 0.6  \\
kernel regularizer & L2 with (0.01)  \\
\hline
\end{tabular}
\end{center}
\caption{\label{parameter} Proposed Models (KEIS-BiGRUCNN) Hyper-parameters }
\end{table}

\subsubsection{BERT Fine-Tuned (BERT-Bi)}
\label{BERT:layout}
In the recent years, contextual embedding shows the significant progress in the NLP research field. Consequently, according to the reported results in several researches ({\em i.e.} \newcite{zhu-etal-2019-um}) it's it's outperform the deep learning approaches. The transformer considered as an encoder-decoder architecture applied on attention mechanisms tasks. More particularly, Google has been released BERT \cite{devlin2018bert} which stands for Bidirectional Encoder Representations from Transformers. BERT is a deeply bidirectional architecture which means BERT during the training step can learn the important features from both sides of a word’s context. They applied masked language modeling this means the model depends on the position of the words to infer the information among them. The pre-trained BERT model can be fine-tuned.\\

Our intention to solve the offensive detection shared task using fine-tuned the BERT by adding Gaussian Noise layer followed by bidirectional LSTM \cite{lstm} consist of 300 neurons, GRU \newcite{chung2014empirical} consist of 300 neurons, and global average pooling (GAP) to extract the discriminative features from the past layer and keep them to the next layer. L2 regularization and Dropout have been used to prevent overfitting. The classification layer used to find the final predictions dense layer of 1 neuron with sigmoid activation function and TruncatedNormal kernal initializers. Consequently, the implemented approach called BERT-Bi. \\

\begin{figure*}
\centering
  \includegraphics[width=1\textwidth]{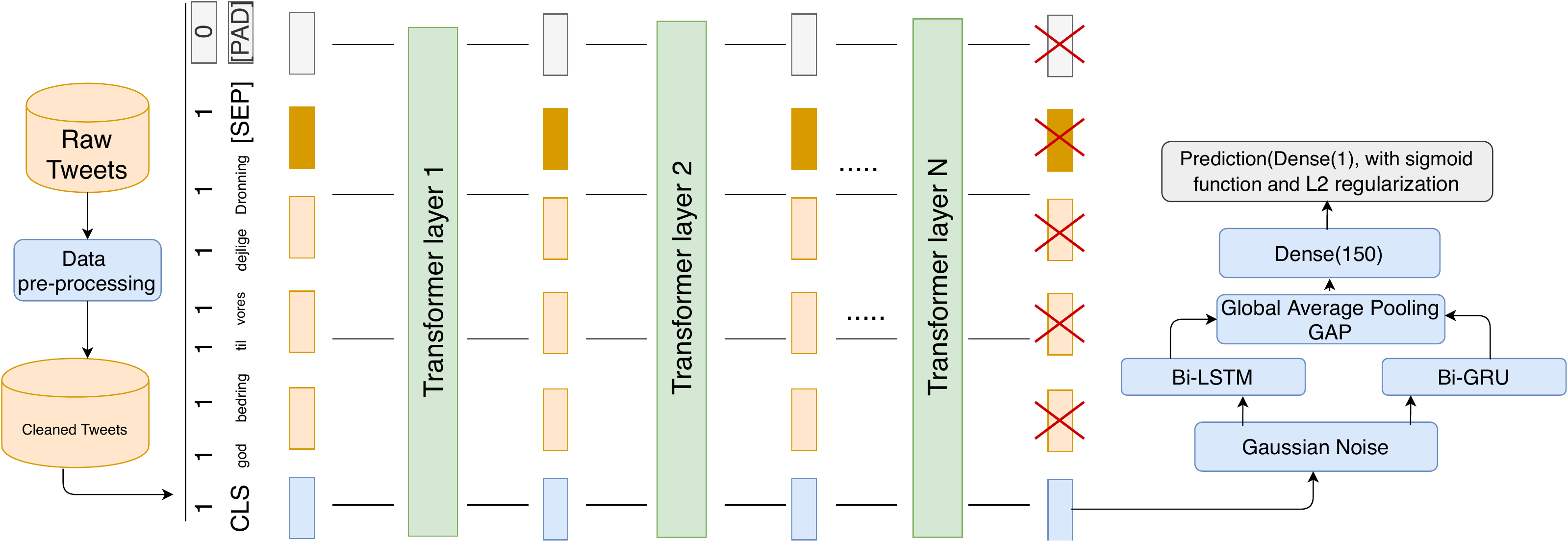}
  \caption{The architecture of BERT fine-tuning model (BERT-Bi)}
  \label{fig:BERT2}
\end{figure*}

The implemented BERT-Bi based on transfer learning architecture that has used in common specially in image classification and computer vision \cite{litjens2017survey}. Moreover as we mentioned earlier in Sec \ref{related:pdf}, the applied of transformers show the promising results compared to deep learning approaches. For instance, BERT developers created several pre-trained models such as uncased, cased, and multi\_cased to represent the semantic relationships among text as well as it could be applied as an independent classifier in different NLP domains ({\em i.e.} offensive language detection). In this research, we used multi\_cased model since it trained on multi languages based on transfer learning architecture to tackle the shared task problem. The BERT-Bi architecture shown in Fig. \ref{fig:BERT2} used to solve sub-task A for three languages namely: Greek, Danish, and Turkish. Whereas the special [CLS] should be added at the beginning of each tokinzed tweet. the special [SEP] should be added at the end of each tokinzed tweet. The  the Attention Mask represented as an array of 1's and 0's. In order to implement proposed model, several parameters have been used. According to the experimental results the best parameter as follows: batch size= 16, optimizer= Adam, learning rate= 2e-5, and finally BERT max length= 60.

\section{Experimental Setup}
\subsection{Dataset}
Multilingual dataset have been provided with five languages namely: Arabic, Danish, English, Greek, and Turkish. The annotation follows the hierarchical tagset for the prevuos Offensive Language Identification Dataset (OLID) \newcite{OLID}. The provided dataset to tackle Task-12 at SemEval 2020 which has been obtained from Twitter using API's. Task 12 offensEval 2020 provided three sub-tasks: (1) if the tweet offensive (OFF) or non-offensive (NOT), (2) if the tweet is targeted (TIN) or un-targeted (UNT), and (3) If the target is an individual (IND), group (GRP) or other (OTH). The provided dataset is multilingual and imbalanced refers to the distribution for each sub-task including the labels provided for the three sub-tasks with tab separated file format. Table \ref{dataset_distrbution} shows the distribution of the available dataset. Table \ref{dataset_EXAMPLE} provide examples that represents dataset for all languages.

\begin{table*}[t!]
\centering
\begin{adjustbox}{max width=\textwidth}
\begin{tabular}{lllllll}
  Data File/Lang & Arabic & Danish & Greek & Turkish & Eng-B& Eng-C\\ \hline
  Train & \makecell{OFF= 1410\\ NOT= 5590} &\makecell{OFF= 398\\ NOT= 2568}&\makecell{OFF= 2539\\ NOT= 6260}&\makecell{OFF= 6132\\ NOT= 25630}&\makecell{TIN= 149550\\UNT= 39424}&\makecell{IND= 120330\\ GRP= 22176\\ OTH= 7043}\\
  
  Validation &\makecell{OFF= 179\\ NOT= 821}&\makecell{20\% \\ split}&\makecell{20\% \\ split}&\makecell{20\% \\ split}&\makecell{20\% \\ split}&\makecell{20\% \\ split}\\
  
  Test &\makecell{OFF= 402\\ NOT=1598 }&\makecell{OFF= 41\\ NOT= 288}&\makecell{OFF= 242\\ NOT= 1302}&\makecell{OFF= 716\\ NOT= 2812}&\makecell{UNT= 572\\ TIN= 850}&\makecell{IND= 580\\ GRP= 190\\ OTH= 80}\\
\end{tabular}
\end{adjustbox}
\caption{\label{dataset_distrbution}The Dataset distribution}

\end{table*}

\begin{table*}[t!]
\centering
\begin{tabular}{llll}
  id & lang & tweet & label\\
  \hline
   1689 & Arabic & @USER \textAR{اتعلم يا متعصب يا معدوم الذمه} & OFF\\
   1713 & Danish & Haha, det er genialt! &NOT \\
   19321 & Turkish & @USER Burası da fena değil atkafalı &OFF \\
   1159528564925984768 &English & everyone talks shit in LA & TIN  \\
\end{tabular}
\caption{\label{dataset_EXAMPLE}Examples from different languages that represents the dataset}
\end{table*}

\subsection{Data Pre-processing}
The convenient process regarding social network dataset such that, Facebook and Twitter, tweets, and posts which contain such noisy data and slang language. In the raw text, it should remove the special character, punctuation marks ( *,\^@\#-(|{}), URLs, and user mentions. The normalization was necessary since some words written on short-cut format, the elongation was also removed (e.g congrats \textAR{مبروووووك}). Finally, numbers and English characters were also removed for Arabic. Moreover, the emojis have been removed.

\subsection{Embeddings}
Several well-known word embedding are provided to extract the vector representation of the input tweets with aims to capture the semantic features for each word and the relationship among them. Word2Vec has been provided by \newcite{mikolov2013distributed}, Glove \cite{pennington2014glove}, AraVec \newcite{aravec} and the recent contextual embedding ElMo by \cite{peters2018deep} and BERT \newcite{devlin2018bert}. In this research, we used AraVec, Word2Vec and the pre-trained BERT embedding to trained the performed model. It is a language representation model and becoming the state of art model for the most of NLP research.
\subsection{Data Augmentation}
It is a way to improve the performance of NLP models, data augmentation should appear on a deep understanding of our dataset including structure and content. The impact of using data augmentation technique will depend on that technique itself, where each one able to learn something different compare to others and generate a different impact as well. There are several techniques used in the data augmentation, in this research we well performed the technique depending on pre-trained AraVec \newcite{aravec}. The first step load embedding and prepare the dataset. Afterward, making a synonym dictionary depending on the most frequent words.

\subsection{Discussion}
\begin{figure*} 
\begin{minipage}[t]{.45\linewidth}
\includegraphics[width=\linewidth]{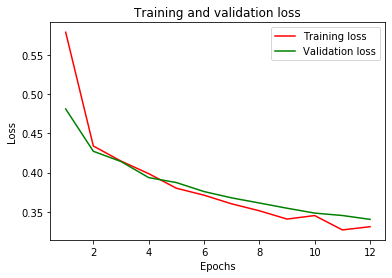}
\caption{KEIS@BiGRU loss plots for training and validation data}
\label{fig:loss_bi}
\end{minipage}\hfill
\begin{minipage}[t]{.45\linewidth}
\includegraphics[width=\linewidth]{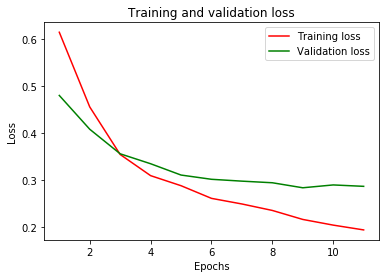}
\caption{KEIS@CNN loss plots for training and validation data}
\label{fig:loss_cnn}
\end{minipage}\hfill
\end{figure*}

Our results extracted using SAJA CODALab user name and the team name is KEIS@JUST. The reported results on the validation set are presented in table \ref{dev_results_arabic} for Arabic and table \ref{result_dev_other} for other languages. Table \ref{dev_results_arabic} presents the results using data augmentation for Arabic language. It shows the enhancement of using the augmentation regarding the overall performance the KEIS-BiGRUCNN approach achieved F1= 87.9\% on Arabic validation set. Moreover, BERT-Bi approach achieved F1= 78\% on Danish validation set (see table \ref{result_dev_other}). As we mentioned above in sec \ref{ssec1:layout} presents KEIS@JUST System to present the results which consist of a)KEIS-BiGRUCNN used to solve sub-task A for Arabic language and sub-task B,C for English language. b) KEIS-BERT-Bi used to solve sub-task A for other languages. To prevent overfitting during the training step, the early stopping and checkpoints have used among the training set and the validation set and keep track of the loss value at the end of each training epoch. Moreover, the learning rate reduction has used. Fig. \ref{fig:loss_bi} and Fig. \ref{fig:loss_cnn} show the model training.

\begin{table}[t!]
\begin{center}
\begin{tabular}{|l|l|l|}
\hline \bf Model & \bf without & \bf with \\ 
\bf  & \bf aug & \bf  aug \\ \hline
KEIS-BiGRU & 83.6\% & 87.6 \\
KEIS-CNN & 83.5\%  & 87.3\\
KEIS-BiGRUCNN & 83.8\% & 87.9 \\
\hline
\end{tabular}
\end{center}
\caption{\label{dev_results_arabic} The results on the validation set for Arabic language (aug refers to augmentation) }
\end{table}

\begin{table*}[t!]
\centering
\begin{tabular}{ll|ll}
  Task A Multilingual using  BERT-Bi & F1 & Task B/C English using BiGRUCNN & F1 \\ \hline
   Danish & 78\% & Task B & 48.3\%  \\
   Greek & 78.5\% &Task C&58.9\% \\
   Turkish & 72\% &-&-
\end{tabular}
\caption{\label{result_dev_other}: The results on the validation set for other languages/all results computed using Macro-F1}
\end{table*}

\section{Results and Findings}
\label{secr:results}
In order to evaluate the implemented approaches, F1-Macro has been used according to the shared task instruction. Table \ref{results1} presents the results of the participants models for Arabic, Danish, and Greek languages and macro-average results, as presented in the table, results for are competitive. Moreover, Table \ref{results2} presents the result for Turkish and English language. The KEIS@JUST team, achieved 11th place with F1= 86.55\% in Arabic, achieved 12th place with F1= 76.1\% in Danish, achieved 28th place with F1= 77.3\% in Greek, achieved 32th place with F1= 73.3\% in Turkish. In contrast, achieved F1= 27.7\% in English sub-task B and achieved F1= 48.17\% in English sub-task C.

\begin{table*}[t!]
\centering
\begin{tabular}{ll|ll|ll}
  Arabic Rank & F1 & Danish Rank & F1 & Greek Rank & F1 \\ \hline
   1 & 90.17\% & 1 & 81.2\% & 1 & 85.2\% \\
   2 & 90.15\% & 2 & 80.2\% & 2 & 85.1\% \\
  \bf 11 & \bf 86.55\% &\bf 12 &\bf 76.1\% & \bf 28 & \bf 77.3\% \\
\end{tabular}
\caption{\label{results1}: The results on the test set for Arabic,  Danish, and Greek languages/all results computed using
Macro-F1 / Our results appears in bold}
\end{table*}

\begin{table*}[t!]
\centering
\begin{tabular}{ll|ll|ll}
  Turkish Rank & F1 & Eng Rank Task B & F1 & Eng Rank Task C & F1 \\ \hline
   1 & 81.57\% & 1 & 74.6\%  & 1 & 71.45\% \\
   2 & 81.66\% & 2 & 73.6\% & 2 & 66.99\% \\
  \bf 32 &\bf 73.3\% & \bf43 &\bf27.7\% & \bf 33 & \bf 48.17\% \\
\end{tabular}
\caption{\label{results2}: The results on the test set for Turkish and English language-task B and C/all results computed using
Macro-F1 / Our results appears in bold}
\end{table*}

\section{Conclusion}
In this research, presented the KEIS@JUST participation at SemEval-2020 Task 12 which represents shared task on multilingual offensive language. We have participated in all the provided languages for all subtasks except sub-task-A for the English language. Two main approaches have been developed the first one is performed to tackle both languages Arabic and English, a weighted ensemble consists of Bi-GRU and CNN  followed by Gaussian noise and global pooling layer multiplied by weights to improve the overall performance. The second one performed for other languages, we investigated the main impact of developing a transfer learning approach from BERT transformer beside the recurrent neural networks such as Bi-LSTM and Bi-GRU followed by the global average pooling layer for other languages. Word embedding and contextual embedding have been used as features, moreover, we investigated how data augmentation affect the results using Arabic dataset.

\section*{Acknowledgments}
This research is partially funded by Jordan University of Science and Technology, Research Grant Number: 20170107.

\bibliographystyle{coling}
\bibliography{semeval2020}

\end{document}